\documentclass{article}
\usepackage{spconf,amsmath,graphicx}
\usepackage{hyperref}
\usepackage{breakurl}
\usepackage{placeins}
\usepackage{booktabs}
\usepackage{array}
\usepackage{subcaption}
\usepackage{amssymb}
\usepackage{pifont}
\usepackage{amssymb, bm, mathrsfs}
\usepackage{xcolor, adjustbox, booktabs, multirow, colortbl, array, caption}

\title{Physics-Informed Simulation Framework for Realistic Sonar Image Generation and Statistical Validation}

\name{Kamal Basha S and Athira Nambiar}
\address{Department of Computational Intelligence,\\
SRM Institute of Science and Technology,\\
Chennai, India}

\begin{document}
\maketitle

\begin{abstract}
Synthetic sonar datasets offer a scalable alternative to costly real-world acquisition, yet their utility remains limited by the absence of rigorous quantitative validation. We present \textbf{ACOUSIM} (ACOustic SIMulation and Validation Platform), a physics-informed framework that evaluates the statistical alignment between synthetic and real sonar imagery without relying on generative models. A Gazebo-based environment generates sonar-like images by explicitly controlling seabed texture, illumination-driven shadowing, platform altitude, and noise. Realism is quantified against two public sonar datasets such as SeabedObjects-KLSG-II and Sonar Common Target Detection (SCTD) using global intensity and local texture (LBP) distributions, assessed via Kullback--Leibler divergence, Jensen--Shannon divergence, and Earth Mover's Distance. Results show strong texture alignment (KL\,$<$\,0.07) across all classes, with plane-class intensity alignment outperforming ship-class due to shadow geometry complexity. ACOUSIM establishes a reproducible, distribution-level baseline for sim-to-real sonar evaluation, directly supporting reliable dataset validation for underwater image analysis.
\end{abstract}


\begin{figure*}[t]
\centering
\includegraphics[width=0.8\textwidth]{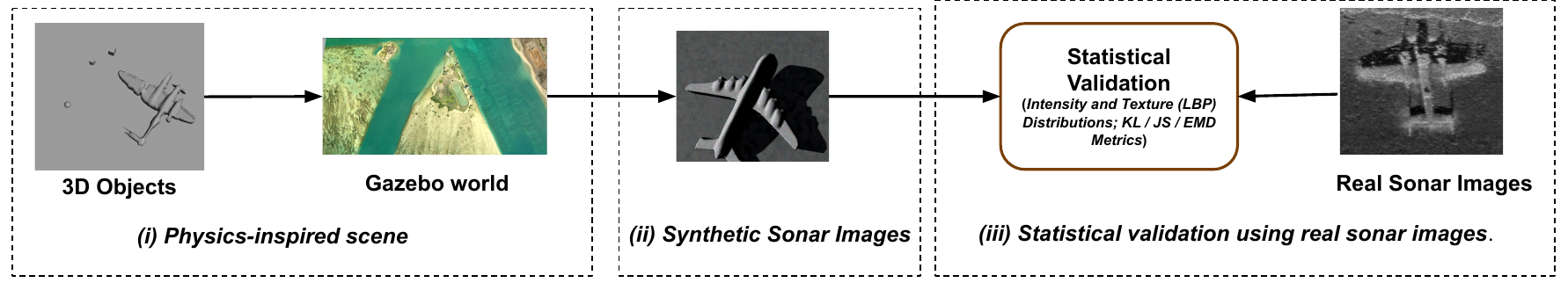}

\caption{High-level overview of \textbf{ACOUSIM}. 3D object models are placed in a Gazebo-based underwater environment to generate synthetic sonar-like images, which are statistically validated against real sonar datasets.}
\label{fig:framework_overview}
\end{figure*}

\vspace{-0.3cm}
\section{Introduction}
\label{sec:intro}

Sonar image analysis plays a critical role in ocean and offshore applications such as seabed mapping, underwater inspection, mine countermeasures, and object detection for autonomous and remotely operated underwater systems~\cite{yang2020underwater}. Both civilian
and defense sectors rely on high-quality sonar imagery to support safe and reliable underwater operations. However, the development and validation of robust perception algorithms remain constrained by the limited availability of diverse and high-fidelity real-world sonar datasets.

The acquisition of real sonar imagery in marine environments
is associated with high operational cost, logistical complexity,
and strict security  restrictions~\cite{williams2010autonomous}. Field experiments require
specialized vessels, trained personnel, and calibrated equipment,
making large-scale and repeatable data collection impractical.
Controlled tank experiments, while comparatively cost-effective,
provide limited environmental variability and do not adequately
capture realistic seabed textures, object–shadow interactions, and
range-dependent intensity variations. As a result, publicly available sonar datasets remain limited in both quantity and diversity,
restricting systematic evaluation of underwater perception systems


To address these limitations, recent studies have explored learning-based sonar image generation techniques, including generative adversarial networks (GANs), diffusion models, and style-transfer approaches~\cite{peng2025multi,ma2025enhancing}. In parallel, simulation-driven frameworks such as S3Simulator~\cite{kamal2024s3simulator}, and other virtual underwater sensing platforms have been developed to generate synthetic sonar imagery under controlled environmental and sensor configurations. While these approaches improve visual realism and dataset scalability, their validation is often limited to qualitative inspection or indirect performance gains in downstream machine learning tasks. Quantitative assessment of how closely synthetic images align with real sonar data distributions remains comparatively underexplored.

However, evaluation almost universally relies on qualitative inspection or indirect downstream metrics, leaving a fundamental question unanswered: \textit{how statistically close is synthetic sonar data to real?} Standard image quality metrics PSNR, SSIM, FID, KID~\cite{wang2004image,heusel2017gans} are calibrated for natural optical images and cannot capture sonar-specific intensity and texture statistics. Mixing real and synthetic data during training further obscures the intrinsic domain gap~\cite{shin2022synthetic,lian2025underwater}.

This paper addresses this gap by presenting ACOUSIM
(ACOustic SIMulation and Validation Platform), a physics consistent simulation and validation framework for sonar image
generation with a specific focus on statistical distribution alignment rather than learning-based performance metrics. In this
work, the term physics-informed refers to explicit control over
acquisition-relevant parameters that influence sonar image appearance, including platform (camera) height in the range of 10–
20 m, roll, pitch, and yaw variations of the illumination source
to simulate object shadowing, and the inclusion of Gaussian and
speckle noise to approximate characteristics of coherent acoustic
imaging. Rather than modeling full acoustic wave propagation,
these parameters are used to reproduce first-order visual effects
observed in real sonar imagery in a controlled and interpretable
manner. To the best of our knowledge, this work represents one
of the first systematic efforts to evaluate synthetic sonar image
realism through direct statistical distribution alignment with real
sonar datasets, without coupling the analysis to learning-based
performance metrics.

Statistical validation is performed using Kullback–Leibler
(KL) divergence~\cite{kullback1951information}, Jensen–Shannon (JS) divergence~\cite{lin1991divergence}, and
Earth Mover’s Distance (EMD)~\cite{rubner2000earth}. By decoupling synthetic data evaluation from machine learning models,
this work establishes a reproducible and interpretable benchmark
for real versus synthetic sonar image comparison. The proposed
framework, referred to as ACOUSIM, provides a foundational
step toward physics-aware dataset validation and supports future
development of digital twins for underwater sensing systems. The
main contributions of this work are summarized as follows:

\begin{itemize}
\setlength{\itemsep}{1pt}\setlength{\parskip}{0pt}
\item \textbf{ACOUSIM}: a physics-informed sonar simulation and standalone statistical validation platform, independent of generative models.
\item \textbf{Statistical validation scheme} based on global intensity and LBP texture distributions, quantified with KL, JS, and EMD metrics.
\item \textbf{Evaluation on two real sonar datasets} (KLSG-II, SCTD) using dataset-level analysis and stratified five-fold robustness assessment.
\end{itemize}




\vspace{-0.4cm}
\section{Related Work}
\label{sec:rw}

The development of sonar image datasets for underwater perception and analysis has been actively explored through three primary directions: (i) Acquisition of real sonar data, (ii) Learning-based sonar image synthesis, and (iii) Simulation-driven synthetic data generation.

\textbf{Real sonar datasets} are commonly acquired using side-scan sonar (SSS) systems deployed in controlled underwater surveys and seabed inspection missions. Existing datasets include shipwreck segmentation benchmarks~\cite{sethuraman2025machine}, semisynthetic training datasets for underwater object classification~\cite{huo2020underwater}, and Sonar Common Target Detection Dataset~\cite{zhang2021self}. These datasets provide annotated sonar imagery for tasks such as target detection, classification, and segmentation, with validation typically performed using expert-labelled ground truth and downstream perception benchmarks. While real-data acquisition provides physically authentic sonar measurements, large-scale collection and annotation remain expensive, time-consuming, and operationally constrained.

\textbf{Learning-based sonar synthesis} approaches attempt to generate realistic sonar imagery from limited real data using generative models. Existing methods include CycleGAN-based sonar translation frameworks~\cite{koo2024cycle}, hybrid ray-tracing with GAN refinement approaches~\cite{li2024side}, multi-view GAN-based sonar generation~\cite{peng2025multi}, and diffusion-model-based sonar enhancement frameworks~\cite{ma2025enhancing}. These approaches have demonstrated improved visual realism and promising downstream recognition performance for tasks such as classification, detection, and segmentation. Evaluation is typically performed using perceptual similarity metrics such as FID and KID together with downstream task-level accuracy measures.

\textbf{Simulation-driven approaches} focus on reproducing sonar image characteristics through analytical, geometric, or physics-inspired acoustic modelling. Unreal Engine-based underwater sonar simulators~\cite{shin2022synthetic}, ray-acoustic modelling frameworks~\cite{lian2025underwater}, and S3Simulator~\cite{kamal2024s3simulator} generate synthetic sonar imagery under controlled environmental and sensor configurations. Validation is commonly performed using target recognition benchmarks, shadow consistency analysis, or limited statistical comparisons against real sonar imagery.

Despite these advances, validation of synthetic sonar datasets across existing literature remains predominantly indirect, relying on perceptual quality metrics or downstream task performance. Systematic validation based on direct statistical distribution alignment between real and synthetic sonar image domains remains comparatively underexplored.
\section{Methodology}
\label{sec:method}

ACOUSIM consists of three stages (Fig.~\ref{fig:framework_overview}): (i) physics-inspired scene modelling, (ii) synthetic image generation with noise, and (iii) standalone statistical validation.

\begin{figure*}[t]
\centering
\includegraphics[width=0.6\textwidth]{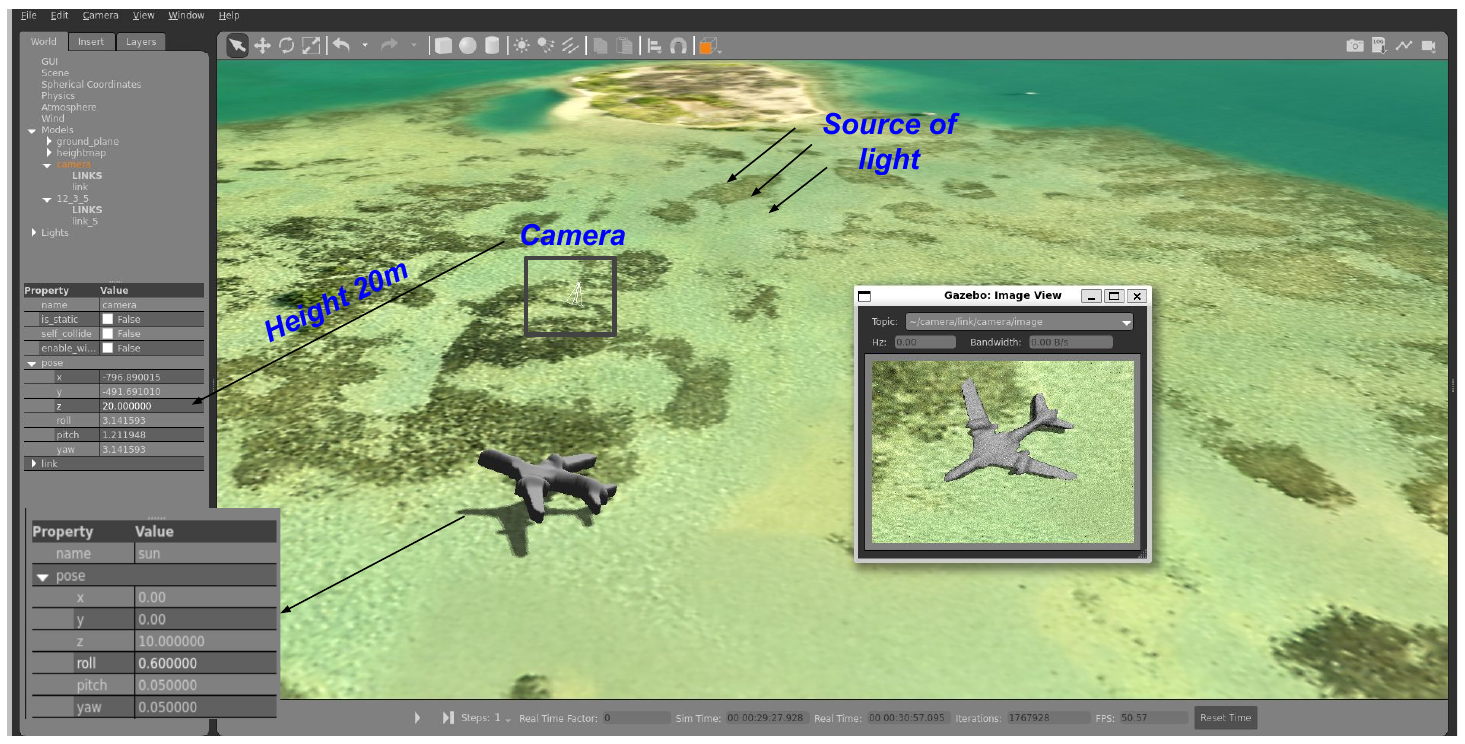}

\caption{Gazebo-based scene configuration in ACOUSIM showing object placement, camera-height variation, and shadow generation through roll, pitch, and yaw adjustment of the illumination source.}
\label{fig:gazebo_world}
\end{figure*}

\subsection{Physics-Inspired Scene Modelling}

Synthetic sonar-like images are generated in a Gazebo environment~\cite{koenig2004design} by emulating first-order visual characteristics of sonar imagery like object shadows, seabed backscatter, and range-dependent intensity, rather than modelling full acoustic propagation. Planar seabed surfaces with varied textures simulate heterogeneous backscatter; 3D object models introduce shadow and highlight regions.

Shadow geometry is controlled via directional illumination and viewpoint orientation (roll, pitch, yaw), enabling diverse shadow patterns without altering object structure. Platform altitude is varied between 10--20\,m. Image formation is approximated as:
\begin{equation}
I(x,y) = L(x,y)\cdot R(x,y),
\label{eq:formation}
\end{equation}
where $L(x,y)$ encodes illumination and shadowing effects, and $R(x,y)$ denotes surface reflectance. This first-order model is sufficient for distribution-level statistical comparison.

\subsection{Noise Modelling and Image Processing}

Simulation outputs are extracted from video sequences recorded under both static and dynamic conditions. To improve realism, synthetic sonar images are degraded using additive Gaussian noise and multiplicative speckle noise, which approximate background disturbances and coherent acoustic scattering effects commonly observed in sonar imagery. The resulting noisy image is formulated as:
\begin{equation}
I_S(x,y) = \bigl(I(x,y) + \mathcal{N}(0,\sigma^2)\bigr)\cdot S(x,y).
\label{eq:noise}
\end{equation}
where \(I(x,y)\) denotes the clean synthetic image and \(I_S(x,y)\) represents the final noise-degraded sonar image after Gaussian and speckle noise modelling. $\mathcal{N}(0,\sigma^2)$ represents additive Gaussian noise and $S(x,y)$ denotes multiplicative speckle noise. Finally, all images are converted to grayscale, and gradient-magnitude mapping is applied to enhance structural features prior to statistical analysis.

\subsection{Statistical Validation and Subset Selection}
\noindent\textbf{Physics- and acquisition-consistent subset selection:}
Synthetic sonar images generated in Gazebo often contain large seabed regions and wider scene coverage compared to real sonar datasets, where targets are typically cropped around the object region as shown in Figure~\ref{fig:qualitative}. To ensure acquisition-consistent comparison, synthetic samples are spatially cropped and zoom-adjusted to match the object scale, framing, and viewing characteristics observed in real sonar imagery. All images are converted to grayscale, resized to \(256\times256\), and normalized to an 8-bit intensity range before statistical validation.


Global intensity and LBP texture histograms are constructed for real ($P$) and synthetic ($Q$) distributions. Distribution similarity is quantified using three complementary metrics. KL divergence measures directional information loss:
\begin{equation}
D_{\mathrm{KL}}(P \| Q) = \sum_{i} P(i) \log \frac{P(i)}{Q(i)},
\label{eq:kl}
\end{equation}
Jensen--Shannon divergence provides a symmetric, bounded measure via mean distribution $M = \tfrac{1}{2}(P+Q)$:
\begin{equation}
D_{\mathrm{JS}}(P,Q) = \frac{1}{2}D_{\mathrm{KL}}(P\|M) + \frac{1}{2}D_{\mathrm{KL}}(Q\|M),
\label{eq:js}
\end{equation}
and Earth Mover's Distance (EMD) quantifies the minimum transport cost between distributions:
\begin{equation}
\mathrm{EMD}(P,Q) = \inf_{\gamma \in \Gamma(P,Q)} \mathbb{E}_{(x,y)\sim\gamma}[|x - y|],
\label{eq:emd}
\end{equation}
where $\Gamma(P,Q)$ is the set of admissible transport plans. EMD is particularly sensitive to distribution shifts caused by shadow displacement and illumination variation.

To interpret metric magnitudes consistently, alignment is categorised as: \textit{High} ($D < 0.2$), \textit{Moderate} ($0.2 \leq D < 0.7$), and \textit{Low} ($D \geq 0.7$), applied to KL divergence values. A stratified five-fold is applied for robustness assessment, reporting mean $\pm$ standard deviation across folds.

\vspace{-0.1cm}
\section{Experimental Results}
\label{sec:results}

\subsection{Qualitative Result}

\begin{figure}[t]
\centering

\textbf{Before physics-consistent subsetting} \\[1mm]

\begin{minipage}{0.38\linewidth}
\centering
(a) Plane class \\[0.5mm]
\includegraphics[width=0.48\linewidth]{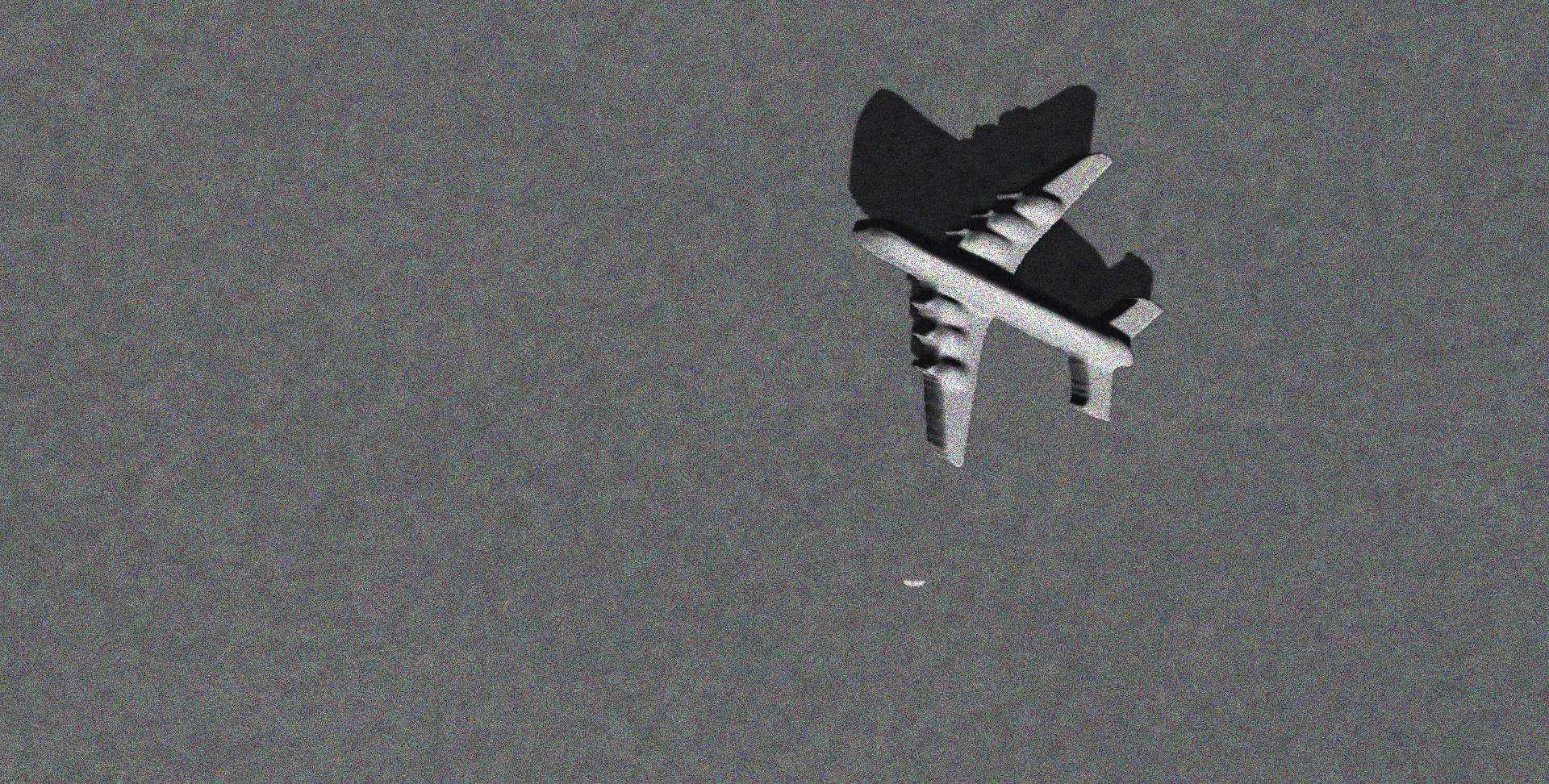}
\includegraphics[width=0.48\linewidth]{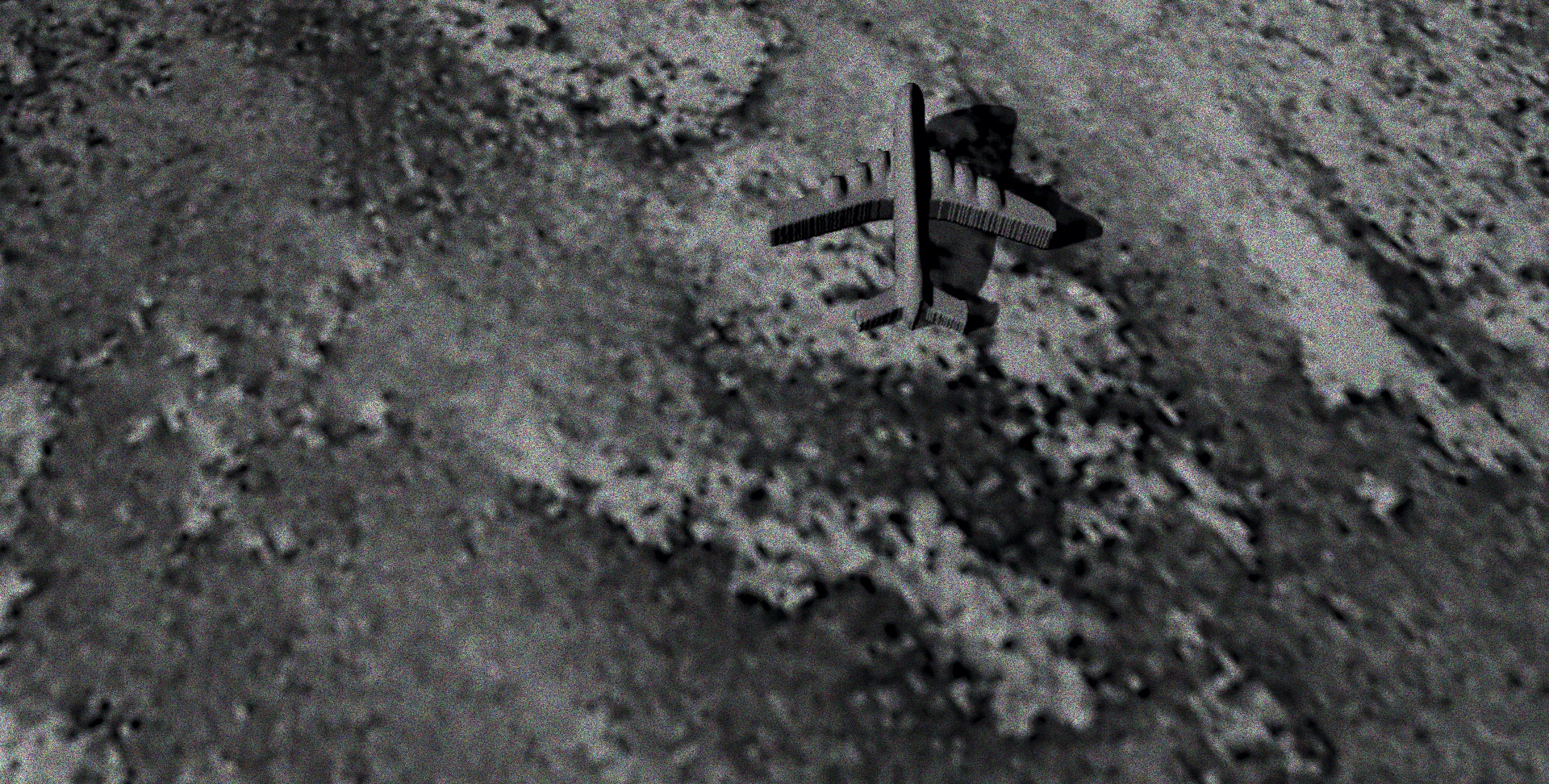}
\end{minipage}
\hspace{2mm}
\begin{minipage}{0.38\linewidth}
\centering
(b) Ship class \\[0.5mm]
\includegraphics[width=0.48\linewidth]{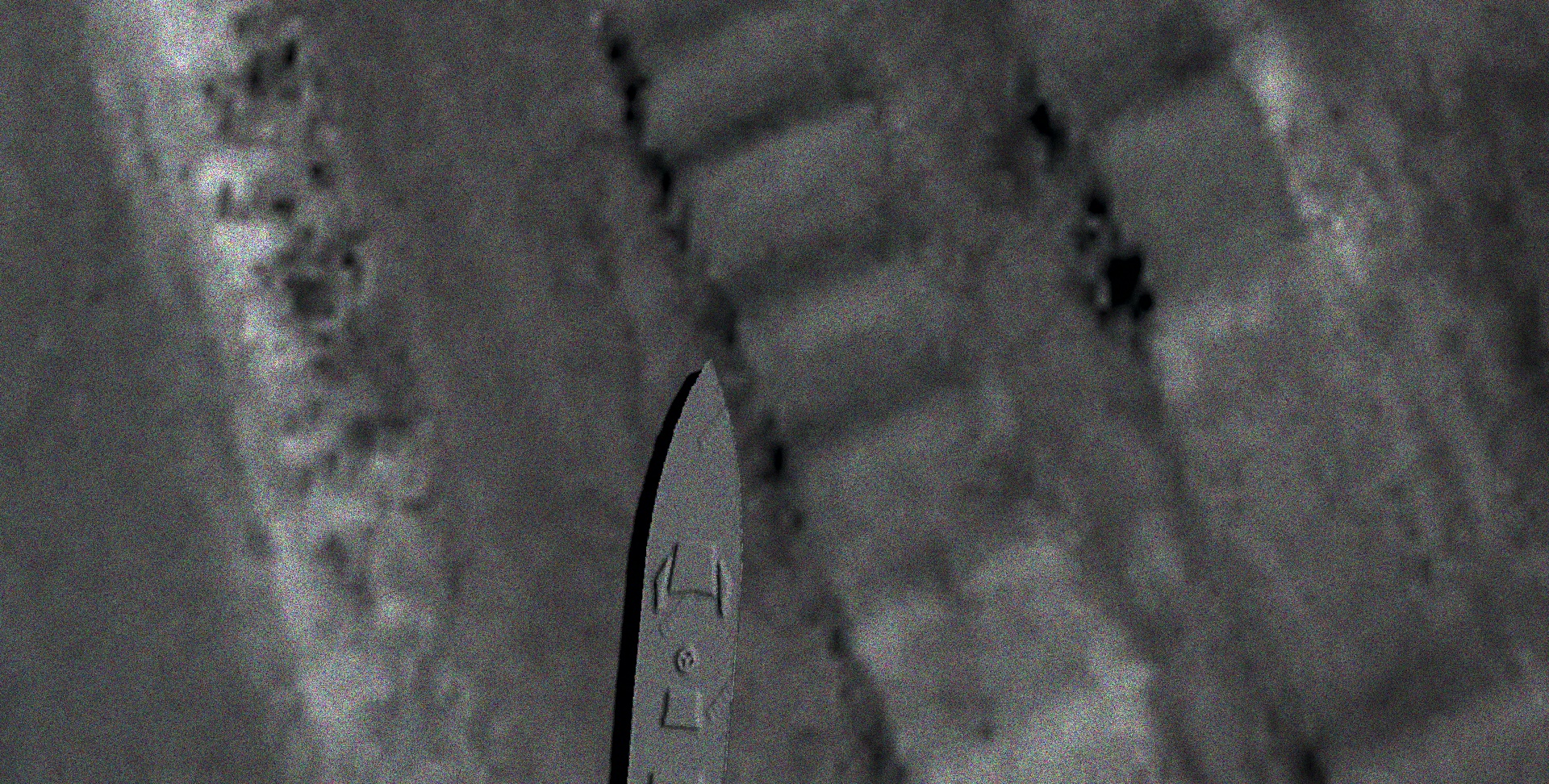}
\includegraphics[width=0.48\linewidth]{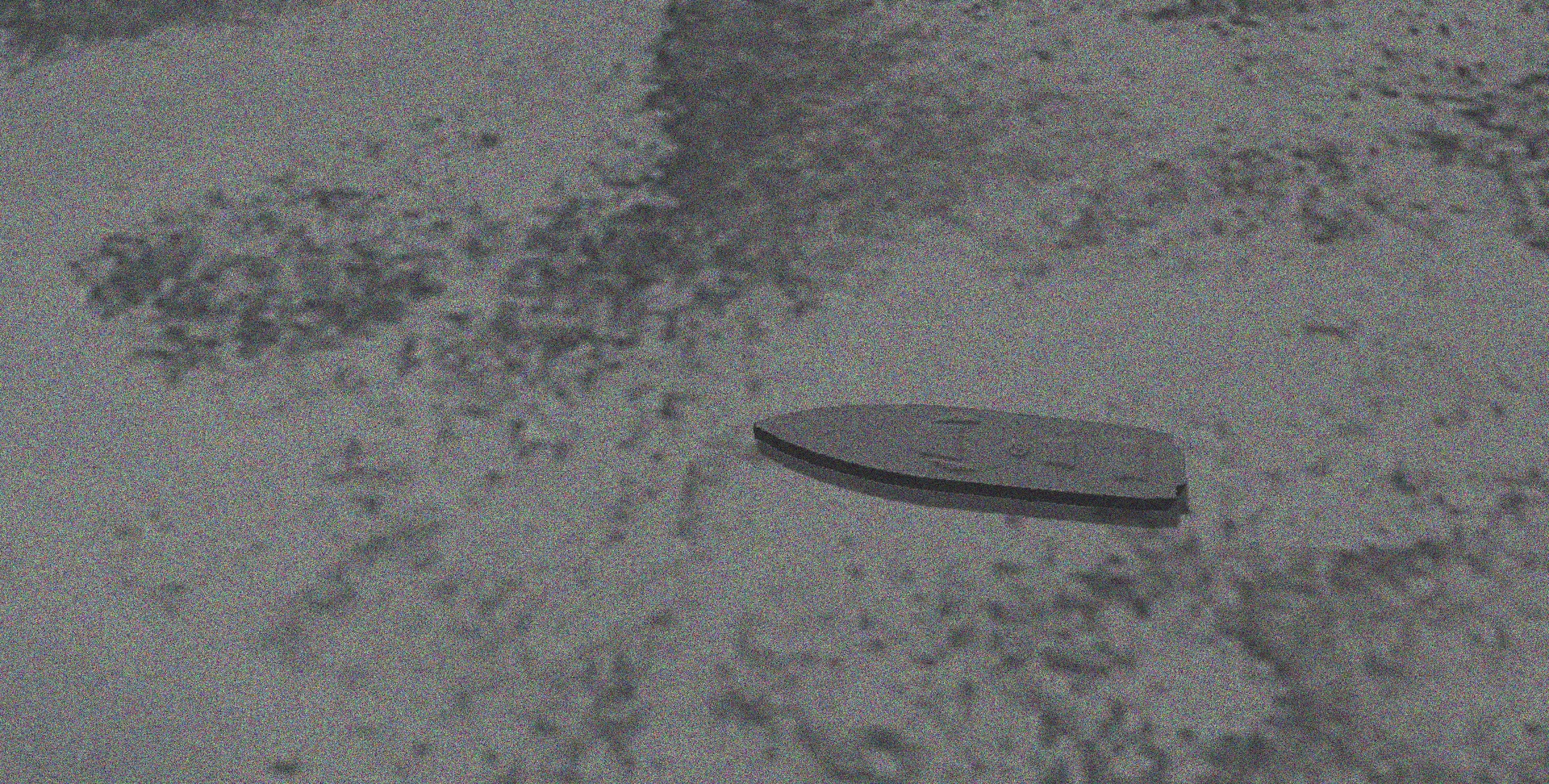}
\end{minipage}

\vspace{2mm}

\textbf{After physics-consistent subsetting} \\[1mm]

\begin{minipage}{0.38\linewidth}
\centering
(c) Plane class \\[0.5mm]
\includegraphics[width=0.48\linewidth]{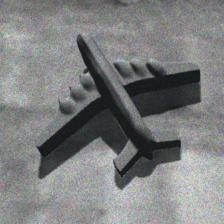}
\includegraphics[width=0.48\linewidth]{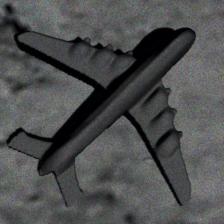}
\end{minipage}
\hspace{2mm}
\begin{minipage}{0.38\linewidth}
\centering
(d) Ship class \\[0.5mm]
\includegraphics[width=0.48\linewidth]{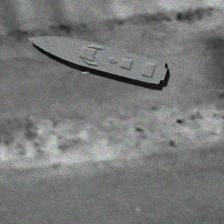}
\includegraphics[width=0.48\linewidth]{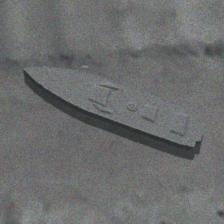}
\end{minipage}

\caption{Qualitative comparison of synthetic sonar samples before and after physics-consistent subsetting. (a) Plane-class samples before subsetting, (b) ship-class samples before subsetting, (c) plane-class samples after subsetting, and (d) ship-class samples after subsetting. The subsetting process improves structural consistency in object boundaries and shadow regions by suppressing excessive seabed background variability while preserving inter-sample diversity.}
\label{fig:qualitative}
\end{figure}





Physics-consistent subsetting markedly improves shadow geometry and boundary consistency across both classes, as shown in Fig.~\ref{fig:qualitative}. Samples (a) and (c) correspond to the plane class before and after subsetting, respectively, while (b) and (d) represent the corresponding ship-class samples. The before-and-after comparison demonstrates that the subsetting process effectively crops the region of interest for validation by excluding excessive seabed background regions while preserving the object profile and associated shadow structure. This indicates that unfiltered synthetic samples contain substantial background variability that does not accurately reflect real sonar acquisition conditions.

\subsection{Quantitative Result}

Quantitative evaluation is performed using KL divergence, JS divergence, and EMD to measure the statistical alignment between real and synthetic sonar image distributions. Table~\ref{tab:intensity} presents the dataset-level comparison of global intensity distributions, while Table~\ref{tab:texture} reports the alignment of local texture distributions computed using LBP features.

\subsubsection{Statistical Distribution Alignment}

\begin{table}[t]
\caption{Dataset-level global intensity alignment (real vs.\ S3Simulator synthetic).}
\label{tab:intensity}
\centering
\small
\begin{tabular}{llccc}
\toprule
\textbf{Dataset} & \textbf{Class} & \textbf{KL} & \textbf{JS} & \textbf{EMD} \\
\midrule
KLSG-II & Plane & 0.1804 & 0.0399 & 0.00053 \\
KLSG-II & Ship  & 0.5442 & 0.0844 & 0.00205 \\
SCTD    & Plane & 0.2500 & 0.0571 & 0.00082 \\
SCTD    & Ship  & 0.9173 & 0.1303 & 0.00192 \\
\bottomrule
\end{tabular}
\vspace{-0.1cm}
\end{table}

\noindent\textbf{Dataset-level global intensity alignment:} Strong intensity alignment is observed from Table~\ref{tab:intensity} for the plane class across both datasets, with KL divergence values of 0.1804 for KLSG-II and 0.2500 for SCTD, indicating \textit{Moderate} distribution similarity. In contrast, the ship class exhibits substantially higher divergence, particularly for SCTD (KL,=,0.9173), corresponding to \textit{Low} alignment. This behaviour is primarily attributed to the elongated hull geometry and extended shadow regions of ship targets, which introduce complex intensity variations that are not fully reproduced by the simulation framework.

\begin{table}[t]
\caption{Dataset-level local texture (LBP) alignment (real vs.\ S3Simulator synthetic).}
\label{tab:texture}
\centering
\small
\begin{tabular}{llccc}
\toprule
\textbf{Dataset} & \textbf{Class} & \textbf{KL} & \textbf{JS} & \textbf{EMD} \\
\midrule
KLSG-II & Plane & 0.0679 & 0.0158 & 0.0164 \\
KLSG-II & Ship  & 0.0290 & 0.0070 & 0.0114 \\
SCTD    & Plane & 0.0405 & 0.0111 & 0.0080 \\
SCTD    & Ship  & 0.0359 & 0.0098 & 0.0092 \\
\bottomrule
\end{tabular}
\vspace{-0.1cm}
\end{table}

\noindent\textbf{Dataset-level local texture alignment:}
Consistently strong local texture alignment is observed across all classes and datasets (Table~\ref{tab:texture}), with KL divergence remaining below 0.07 and EMD below 0.016. These results indicate that the simulation effectively preserves local structural characteristics, including micro-texture patterns and spatial gradient distributions, despite variations in global intensity alignment.

\begin{figure*}[t]
\centering

\includegraphics[width=0.38\textwidth]{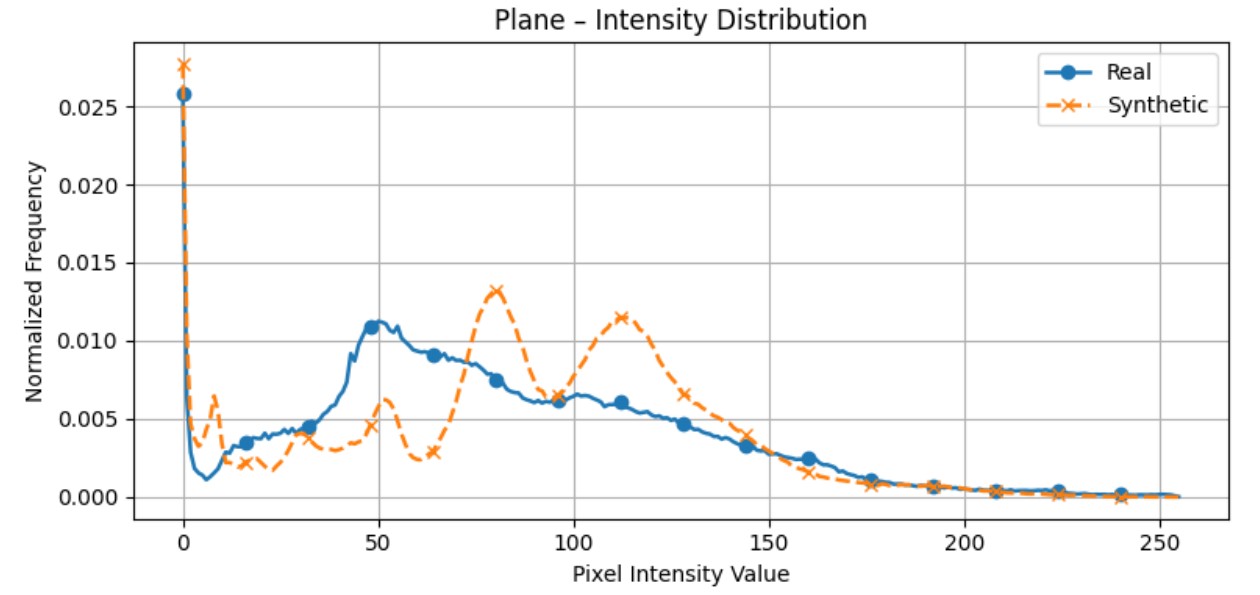}
\includegraphics[width=0.32\textwidth]{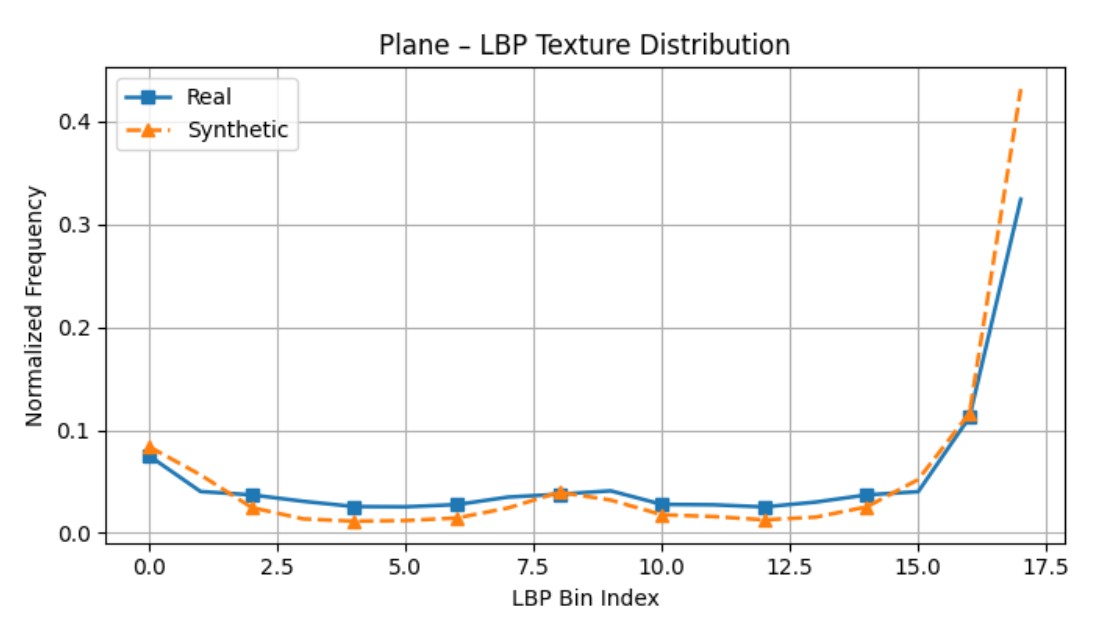}

\vspace{0.5mm}

\includegraphics[width=0.38\textwidth]{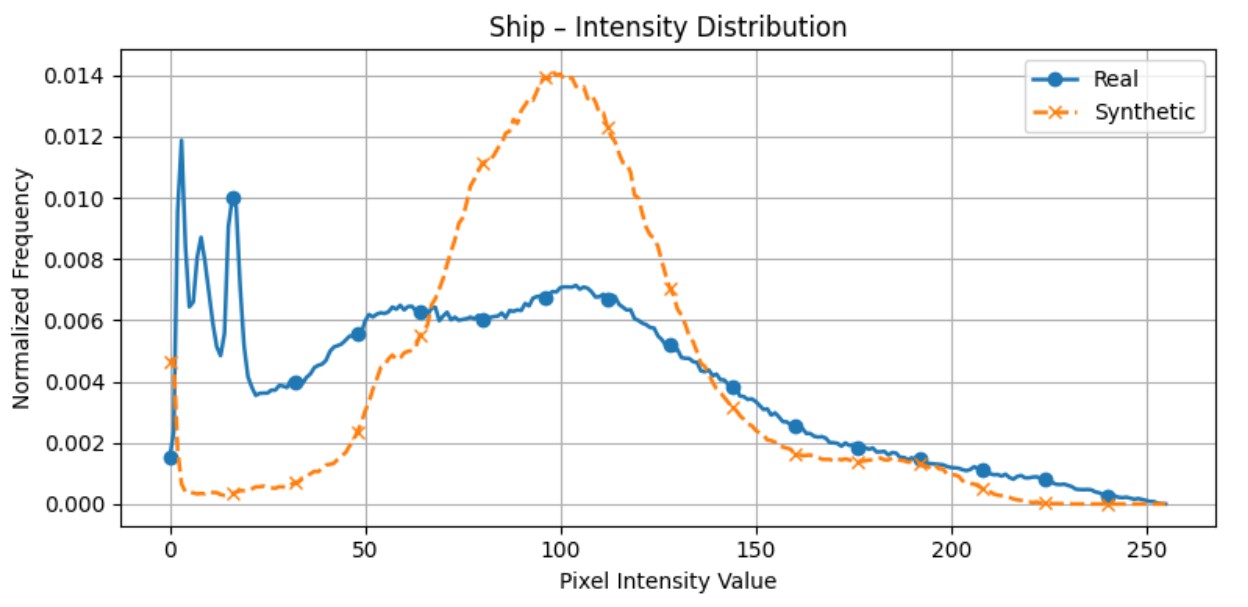}
\includegraphics[width=0.32\textwidth]{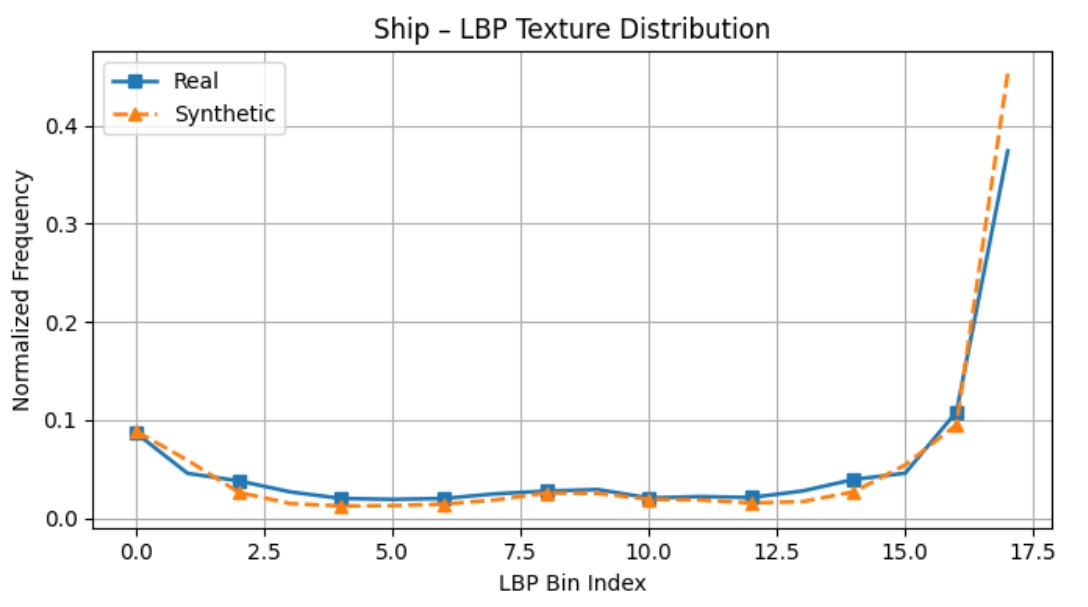}

\caption{Global intensity (left) and LBP texture (right) distribution comparison between real KLSG-II sonar images and S3Simulator synthetic images for plane (top row) and ship (bottom row). Solid curves denote real distributions; dashed curves denote synthetic distributions.}
\label{fig:klsg_dist}
\end{figure*}

\noindent\textbf{Stratified five-fold robustness analysis:} Stratified five-fold robustness results are presented in Table~\ref{tab:kfold_intensity}. For the plane class, low mean divergence with moderate variance confirms stable global intensity alignment across all data splits. For the ship class, higher and more variable divergence (KL\,=\,$1.019\!\pm\!0.231$ for SCTD) reflects sensitivity to viewpoint and shadow geometry variations. Texture KL divergence remains below 0.07 with variance below 0.01 in all folds, confirming that local structural alignment is robust under stratified sampling.


\noindent\textbf{Distribution-level visualization analysis:}
The distribution-level comparison between real and synthetic sonar images for the KLSG-II dataset using both global intensity histograms and LBP-based local texture histograms is depicted in Figure~\ref{fig:klsg_dist}. The top row corresponds to the plane class, while the bottom row represents the ship class. In each plot, solid curves denote real sonar distributions and dashed curves represent synthetic distributions generated by the proposed framework. 
For the plane class, substantial overlap is observed in the mid-intensity range (approximately 60--120), indicating strong global intensity consistency between real and synthetic samples. In contrast, ship targets exhibit a noticeable shift toward higher intensity bins, consistent with stronger shadowing effects and elongated object structures. The LBP distributions show closely aligned peak locations and similar bin magnitudes across both classes, confirming that the simulation effectively preserves local texture characteristics and structural patterns.

\begin{table}[t]
\caption{Stratified five-fold intensity distribution metrics (mean\,$\pm$\,std) comparing real sonar datasets against S3Simulator synthetic data.}
\label{tab:kfold_intensity}
\centering
\small
\begin{tabular}{llcc}
\toprule
\textbf{Dataset} & \textbf{Class} & \textbf{KL} & \textbf{JS} \\
\midrule
KLSG-II & Plane & $0.216 \pm 0.066$ & $0.047 \pm 0.013$ \\
KLSG-II & Ship  & $0.623 \pm 0.106$ & $0.091 \pm 0.010$ \\
SCTD    & Plane & $0.285 \pm 0.040$ & $0.063 \pm 0.010$ \\
SCTD    & Ship  & $1.019 \pm 0.231$ & $0.138 \pm 0.020$ \\
\bottomrule
\end{tabular}
\vspace{-0.1cm}
\end{table}

\vspace{-0.1cm}
\section{Comparison with Existing Approaches}
\label{sec:comparison}

\begin{table}[t]
\caption{Validation strategy comparison across sonar simulation and synthesis works.}
\label{tab:sota}
\centering
\footnotesize
\setlength{\tabcolsep}{5pt} 
\renewcommand{\arraystretch}{1.05}

\begin{tabular}{lcccc}
\toprule
\textbf{Work} & \textbf{Synth.} & \textbf{Phys.} & \textbf{Stat.} & \textbf{Indep.} \\
              &                 &                & \textbf{Eval.} & \textbf{Comp.} \\
\midrule
Shin~\textit{et al.}~\cite{shin2022synthetic}     & \checkmark & \checkmark & $\times$ & $\times$ \\
Koo~\textit{et al.}~\cite{koo2024cycle}           & \checkmark & $\times$   & $\times$ & $\times$ \\
Li~\textit{et al.}~\cite{li2024side}              & \checkmark & $\times$   & $\times$ & $\times$ \\
Peng~\textit{et al.}~\cite{peng2025multi}         & \checkmark & $\times$   & $\times$ & $\times$ \\
Lian~\textit{et al.}~\cite{lian2025underwater}    & \checkmark & \checkmark & $\times$ & $\times$ \\
S3Simulator~\cite{kamal2024s3simulator}           & \checkmark & \checkmark & $\times$ & $\times$ \\
\textbf{ACOUSIM (Ours)}                          & \checkmark & \checkmark & \checkmark & \checkmark \\
\bottomrule
\end{tabular}
\vspace{-0.1cm}
\end{table}
ACOUSIM is compared with prior sonar simulation and synthesis frameworks in Table~\ref{tab:sota}. Existing approaches, including Shin et al.~\cite{shin2022synthetic}, Lian et al.~\cite{lian2025underwater}, and S3Simulator~\cite{kamal2024s3simulator}, incorporate physics-based simulation for synthetic sonar generation but primarily evaluate realism indirectly through downstream recognition performance or qualitative assessment. In contrast, ACOUSIM performs standalone distribution-level statistical validation using KL, JS, and EMD metrics with independent real-versus-synthetic comparison. This enables direct and interpretable quantification of the sim-to-real domain gap without coupling evaluation to any learning model.
\vspace{-0.1cm}
\section{Conclusion}
\label{sec:conclusion}

We presented \textbf{ACOUSIM}, a physics-informed sonar simulation and validation framework that quantifies the real-versus-synthetic domain gap through direct statistical distribution analysis, without generative models or task-dependent evaluation. Across two real sonar datasets, synthetic images show strong local texture alignment (KL\,$<$\,0.07) and moderate-to-strong intensity alignment for plane-class targets, while ship-class divergence highlights challenges in complex shadow geometry. The framework establishes a transparent benchmark for synthetic sonar realism assessment, enabling principled dataset validation before downstream training for underwater image analysis tasks such as detection and segmentation. Future work will extend object and seabed diversity toward a physics-grounded digital twin for underwater sensing.

\bibliographystyle{IEEEbib}
\bibliography{refs}

\end{document}